\documentclass[10pt,conference]{IEEEtran}
\IEEEoverridecommandlockouts
\usepackage{cite}
\usepackage{amsmath,amssymb,amsfonts}
\usepackage{arydshln}
\usepackage{algorithmic}
\usepackage{graphicx}
\usepackage{textcomp}
\usepackage{xcolor}
\usepackage{xcolor,colortbl}
\definecolor{Gray}{rgb}{0.5,0.5,0.5}
\definecolor{gray(x11gray)}{rgb}{0.75, 0.75, 0.75}
\definecolor{honeydew}{rgb}{0.94, 1.0, 0.94}
\definecolor{ivory}{rgb}{1.0, 1.0, 0.94}
\definecolor{linen}{rgb}{0.98, 0.94, 0.9}
\definecolor{magnolia}{rgb}{0.97, 0.96, 1.0}
\definecolor{mintcream}{rgb}{0.96, 1.0, 0.98}
\definecolor{seashell}{rgb}{1.0, 0.96, 0.93}
\definecolor{snow}{rgb}{1.0, 0.98, 0.98}
\usepackage{booktabs}
\usepackage{subcaption}
\usepackage{multirow}
\usepackage{pifont}
\newcommand{\xmark}{\ding{55}}

\newcolumntype{a}{>{\columncolor{Gray}}c}
\def\BibTeX{{\rm B\kern-.05em{\sc i\kern-.025em b}\kern-.08em
    T\kern-.1667em\lower.7ex\hbox{E}\kern-.125emX}}
\begin{document}

\title{Intent Assurance using LLMs guided by Intent Drift}
\makeatletter
\newcommand{\linebreakand}{
  \end{@IEEEauthorhalign}
  \hfill\mbox{}\par
  \mbox{}\hfill\begin{@IEEEauthorhalign}
}
\makeatother

\author{\IEEEauthorblockN{Kristina Dzeparoska}
\IEEEauthorblockA{\textit{Department of Electrical and Computer Engineering} \\
\textit{University of Toronto}\\
Toronto, Ontario \\
kristina.dzeparoska@mail.utoronto.ca}
\and
\IEEEauthorblockN{Ali Tizghadam}
\IEEEauthorblockA{\textit{Department of Electrical and Computer Engineering} \\
\textit{University of Toronto}\\
Toronto, Ontario \\
ali.tizghadam@utoronto.ca}
\linebreakand
\IEEEauthorblockN{Alberto Leon-Garcia}
\IEEEauthorblockA{\textit{Department of Electrical and Computer Engineering} \\
\textit{University of Toronto}\\
Toronto, Ontario \\
alberto.leongarcia@utoronto.ca}
}

\maketitle

\begin{abstract}
Intent-Based Networking (IBN) presents a paradigm shift for network management, by promising to align intents and business objectives with network operations--in an automated manner. However, its practical realization is challenging: 1) processing intents, i.e., translate, decompose and identify the logic to fulfill the intent, and 2) intent conformance, that is, considering dynamic networks, the logic should be adequately adapted to assure intents. To address the latter, intent assurance is tasked with continuous verification and validation, including taking the necessary actions to align the operational and target states. In this paper, we define an assurance framework that allows us to detect and act when intent drift occurs. To do so, we leverage AI-driven policies, generated by Large Language Models (LLMs) which can quickly learn the necessary in-context requirements, and  assist with the fulfillment and assurance of intents.  
\end{abstract}


\section{Introduction}\label{intro}
Intent-based networking (IBN) seeks to align the network behaviour according to business objectives in an automated manner. To do so, IBN must deliver the target behavior based on the intent and must have the ability to consistently ensure that the operational state meets the target state. These main functionalities are referred to as fulfillment and assurance, RFC 9315 \cite{rfc9315}. These two pillars of IBN ensure that network operations are not only effective, but also resilient and adaptive to changing conditions and requirements--allowing for a transformative shift from traditional network management to intent-based management with minimal human intervention. 

The essence of IBN is grounded in its modeling—there needs to be a way to formalize intents, and derive the logic that meets the objectives, resulting with a set of actions to fulfill and assure intents. To capture the requirements at different levels of abstractions in terms of the hierarchy, IBN relies on abstract models and formalisms such as a policy-based language. Policies can provide explanatory IBN, as the actions related to fulfillment and assurance are policy-driven. 

Dynamic changes in network conditions, configurations, including evolving user demands, and scheduled maintenance tasks, can all contribute to the gradual divergence of the network's operational state from its target state. This not only undermines consistency and compliance in terms of the intent's objective, but also poses reliability and performance issues. This deviation is known as the ``intent drift", and can be used as guidance to determine the measures that can realign the network to its target state.

In prior work, we defined \textit{``Emergence"}, a system for intent fulfillment, with elements that facilitate the intent's processing. These include: a policy abstraction that allows us to formalize intents into policy trees, which we execute using closed control loops with finite state machines (such that the transitions are guided by the policies). And, to automatically derive the policy trees, we leverage an LLM pipeline that builds the logic to fulfill intents by generating the necessary policies. In this work, we extend our system with an assurance methodology framework to extract the key performance indicators (KPIs), detect and measure intent drift and extend the LLM pipeline to assure intents. This enables us to close the loop and maintain the intent continuously during its life-cycle.

Our contributions are as follows:
\begin{itemize}
    \item Define a formal intent assurance framework.
    \item Classify and quantify the deviation between target and operational KPIs, to detect intent drift.
    \item Leverage generative models (LLMs) to fulfill and assure intents when intent drift occurs.
\end{itemize}

\section{Background and Related Work}\label{background}
Intent assurance in networking ensures that the network's operational state continuously aligns with its defined intentions by utilizing real-time monitoring, predictive analytics, and AI/ML. To do so, \textit{intent drift} (where network conditions and behaviors deviate from the original intent specifications over time) should be detected and resolved by taking adequate actions in response to the drift, thereby bringing the network back to compliance \cite{rfc9315}. For example, in \cite{martini2023intent} a monitoring and validation component serves to continuously evaluate network slices and trigger corrective actions. 

ML and AI play important roles in IBN, from natural language processing to inference, including predicting when deviations will occur based on past data, and recommending actions for assurance. For example, \cite{zheng2021network} uses time series models to forecast surges in CPU usage. The output of such models should be further analyzed, in addition to other data, in order to determine the correct actions (e.g., VM migrations, scaling, energy optimization) and assure intents during their life-cycle. Similar efforts and proposals have been carried out to evaluate other ML/AI models in \cite{falkner2022intent, samdanis2023ai, perepu2022intent, wang2023artificial, de2022machine, gallego2022machine, perepu2022multi}, as well as heuristics--\cite{mohamed2023automatic}, optimization--  \cite{mai2022end}, and rule-- \cite{9844116} based algorithms.

A recent survey \cite{leivadeas2022survey} details current advances, challenges, as well as common IBN design and components. For assurance, the survey classifies the response into reactive and proactive approaches, with example actions such as flow and VM migration, resource scaling, alerting, and failure recovery. Many open assurance challenges remain, such as intent expression, AI for assurance and lack of datasets, flexible and tailored monitoring, standardizations, multi-domain assurance, security, \cite{leivadeas2023autonomous}. One significant challenge is \textit{diversified assurance}, which requires tuning of the assurance process, such that typical information technology (IT) KPIs can be correlated with unrelated and unknown operational technology (OT) KPIs (e.g., considering Industry 4.0). Thus, a formal assurance methodology that can bring together diverse KPIs, and make comprehensive sense of them, is a must.

Our goal is to develop an assurance methodology that can formalize the requirements and KPIs, detect intent drift, and provide guidance to an LLM to determine the corrective actions. This allows us to extend our system ``Emergence" \cite{10418089} with assurance capabilities. While our current approach is reactive, we can enable proactive behaviour by inferring future state based on historical data analysis.

\section{Intent Assurance Modelling}\label{method}
In our study, we provide definitions that allow us to track the state of the intent at any given time for the purpose of intent assurance. Our goal is to formalize and detect \textit{Intent Drift} and then identify the necessary assurance actions that will bring the intent back to its target state. To do so, we:
\begin{itemize}
    \item Formalize intent with Key Performance Indicators (KPIs). 
    \item Define Intent Drift using KPIs and Assure intents by keeping intent drift within target bounds.
\end{itemize}

\subsection{Formalizing intent with KPIs}
We formalize each intent by extracting the entities and Key Performance Indicators (KPIs) that describe the intent's goal(s). This helps us identify the parameters that need to be met for the intent to be deemed healthy. To do so, a common approach is to use a sequence-to-sequence model (e.g., \cite{jacobs2021hey, 9687516}). Transformer-based models are more capable of understanding the context and semantics of the text, which is helpful when dealing with new and unseen intents. However, creating a good model requires a substantial amount of domain-specific data and computational resources. Thus, we leverage pre-trained LLMs, and use their few-shot learning ability which helps the model learn quickly the task at hand.

We formally define $\overrightarrow{K_{I_i}}$ for an intent $I_i$ as a vector of KPIs, whereby each KPI is described as a name and value pair: 
\begin{equation}
\overrightarrow{K_{I_i}}=(k_1:v_1, k_2:v_2, ..., k_m:v_m).
\label{eq:intentkpi}
\end{equation}

\subsubsection{Few-shot learning to formalize intents and extract KPIs}\label{few-shot}
First, we instruct the LLM model to formalize intents into key:value pairs that capture the intent requirements. Then, we instruct the LLM to extract the KPIs from the formalized intent. 
As an example, the few-shot training prompt contains raw and formalized intents: \textit{``Create collectors in Domain West for gathering Netflow data in the domain, such that the collectors have 99.99\% availability"}; \textit{\{``Domain": ``West", ``Task": ``Create collectors", ``Data Type": ``Netflow", ``Availability": ``99.99\%"\}}. The KPI vector is: \textit{\{``Availability": ``99.99\%"\}}.

This intent will serve as the guiding example, and in fact can be broken down into the following requirements: 1) Creating data collectors in a specific domain, 2) Collecting Netflow Data, and 3) Availability. This implies that in order to ensure that the service is healthy and available, we need to keep track of a set of metrics and KPIs for intent assurance. Once the intent is formalized and fulfilled (explained in Section \ref{evaluation-use}), the LLM can evaluate a composite $service\_health$ KPI to determine the state of the intent, and to assess the $availability$ KPI. To evaluate the health, there are several required KPIs as shown later, which allow us to track and capture the effect of VM failure, collector function failure, etc. Our goal is to be able to measure and compare operational and target KPIs in order to realign the intent and provide assurance. 

\subsection{KPI Quantization}
To determine if an operational KPI deviates from its target value, we apply a quantization to categorize the status of the KPI. The operational KPI value ($k_i^o$) is a continuous variable representing the actual measurement data. We apply a 9-ary quantization ($P_9(k_i^o)$) and as needed we can map the 9-ary into a 3-ary quantization ($P_{9\rightarrow3}(k_i^o)$). Table \ref{table:9-to-3-conversion} shows some examples of the quantizations and their mappings.

\begin{table}[h]
\centering
\begin{tabular}{|c|c|c|c|c|c|}
\hline
\rowcolor{mintcream}
\rule{0pt}{2.5ex} \textbf{$k_i^o$} & \textbf{$P_{9}$} & \textbf{9-ary Description} & \textbf{$P_{3}$} & \textbf{3-ary Description} & \textbf{$P_{9\rightarrow3}$} \\ \hline
$10$ & -4 & Very Low & -1 & Critical & -1 \\ \hline
$15$ & -3 & Low & -1 & Critical & -1 \\ \hline
$25$ & -2 & Slightly Low & 0 & Warning & 0 \\ \hline
$35$ & -1 & Slightly $<$ Normal & 0 & Warning & 0 \\ \hline
$40$ & 0 & Normal & 1 & Normal & 1 \\ \hline
$50$ & 0 & Normal & 1 & Normal & 1 \\ \hline
$65$ & 0 & Normal & 1 & Normal & 1 \\ \hline
$70$ & 0 & Normal & 1 & Normal & 1 \\ \hline
$71$ & 1 & Slightly $>$ Normal & 0 & Warning & 0 \\ \hline
$75$ & 1 & Slightly $>$ Normal & 0 & Warning & 0 \\ \hline
$80$ & 2 & Slightly High & 0 & Warning & 0 \\ \hline
$85$ & 3 & High & -1 & Critical & -1 \\ \hline
$90$ & 4 & Very High & -1 & Critical & -1 \\ \hline
\end{tabular}
\caption{Comparison of quantization (9-ary and 3-ary).}
\label{table:9-to-3-conversion}
\end{table}

\subsection{Logical KPI Assessment}
The overall health, efficiency, and performance of a network are typically assessed by aggregating KPIs and examining them holistically. For example, to formulate an overall assessment of the network health, we can apply a policy function ($f$) over our classified KPIs: $H = f(P_9(\overrightarrow{K})).$ One example is Kleene's logic \cite{kleene1938notation}, such that the overall assessment is based on the worst-performing KPI, which corresponds to taking the minimum as shown below. That is, if any KPI is $-1$, the overall assessment will be $-1$, indicating a critical situation. 
\begin{equation}
f(P_{9}(\overrightarrow{K})) = min(P_{9 \rightarrow 3}(k_1), P_{9 \rightarrow 3}(k_2), ..., P_{9 \rightarrow 3}(k_n)).
\label{kpi}
\end{equation}

\subsection{Detecting KPI Deviations and measuring Intent Drift}
The 9-ary and 3-ary quantization functions categorize the state of each KPI based on how it compares to its target, considering specific thresholds. The quantization doesn't directly measure the difference but rather classifies the state (e.g., normal, warning, critical). To measure the difference between operational and target KPIs, we define:
\begin{equation}
   \Delta \overrightarrow{K}=(\overrightarrow{K_O}-\overrightarrow{K_T})=(\delta_1, \delta_2, ..., \delta_n).
    \label{eq:int_state_1}
\end{equation}

To measure the overall discrepancy between the target and operational KPIs, we can calculate the Euclidean distance using the vector of differences ($\Delta \overrightarrow{K}$). The result represents how far off the operational KPIs are from the targets.
\begin{equation}
\text{Distance} = \sqrt{\sum_{i=1}^{n} \delta_i^2}
\label{eq:euclidean}
\end{equation}

The intent drift can be thought of as the change in the vector $\Delta \overrightarrow{K}$ over time. If this drift is significant, it indicates that the operational performance is increasingly diverging from the targets. And, by calculating the gradient of the distance with respect to each KPI, we can determine the direction and magnitude of change required for each operational KPI to approach its target. This gradient can guide resource allocation or other adjustments. The main steps during assurance include:

\begin{enumerate}
    \item assess KPIs using policies (e.g., quantize). 
    \item periodically calculate $\Delta \overrightarrow{K}$.
    \item compute the distance to gauge overall performance.
    \item calculate gradients to guide adjustments.
\end{enumerate}

We use $\Delta \overrightarrow{K}$ to define the following error function:
\begin{equation}
E(\Delta \overrightarrow{K}) = \sum_{i=1}^{n} \delta_i^2
\label{eq:diff}
\end{equation}
where $\delta_i=k_i^o - k_i^t$ for each KPI. The gradient of the error function with respect to each operational KPI is then:
\begin{equation}
\nabla E = \left( \frac{\partial E}{\partial k_{i_1}^o}, \frac{\partial E}{\partial  k_{i_2}^o}, \ldots, \frac{\partial E}{\partial k_{i_n}^o} \right)
\label{eq:intentdrift}
\end{equation}

Given the definition of $E$, the partial derivative with respect to each operational KPI ($k_i^o$) is given by:
\begin{equation}
\frac{\partial E}{\partial k_{i}^o} = 2 \times (k_i^o - k_i^t) 
\label{eq:gradient}
\end{equation}

The vector of the gradients shown in Equation \ref{eq:intentdrift} is the intent drift vector, where each component represents the direction and magnitude of the change needed for a particular KPI. For example, a positive value indicates that the operational value is higher than the target and needs to be reduced, whereas a negative value indicates the opposite. As the gradients capture the intent drift, they can be used to determine the appropriate assurance actions, which can include for example: resource scaling, application-level optimizations, load-balancing, scheduling, updating thresholds policies, etc. In summary, the gradient doesn't just tell us whether to increase or decrease a value; it provides a quantifiable measure of how much of an adjustment is needed. Ideally, we want the intent drift to be a zero vector, meaning we are aligned with the intent, and there is no drift from it.

\section{KPI Assurance Assessment}
KPIs are defined based on their dependencies such as contributing KPIs and metrics, in line with the intent's requirements. In this work, we define the KPIs for our use-case, and plan to expand the portfolio of KPIs to support other intents. 

For our intent, we combine a set of KPIs into a single, composite KPI--the health of the overall service ($k_{H_s}$). This KPI is dependent on additional KPIs such as resource and software health. The service availability KPI ($k_{A_s}$) is dependent on the service health KPI and as such can be derived from $k_{H_s}$. 

\subsection{Resource Health ($h_r$)}\label{targethealth}
We model the health of a resource (e.g., a VM) as a vector of CPU, RAM, and storage utilization, the VM status and the network connection. To obtain the health of the $i^{th}$ resource ($h_{r_i}$), we apply a policy function ($f$)$^*$ to the above vector. One such example can be a policy to quantize and apply Kleene's logic--in which case, if the below returns $1$, then $h_{r_i}=100\%$. 
\begin{equation}
     h_{r_i} = \bigwedge \left( P(u_{\text{CPU}_i}), P(u_{\text{RAM}_i}), P(u_{\text{storage}_i}), s_{\text{net}_i}, s_{\text{r}_i}\right) .
    \label{eq:resourcehealth}    
\end{equation}
The intent drift vector informs us about the deviations of the above parameters from their targets. For example, let's assume a VM has the following target and operational values, expressed in percentages, where targets can have different thresholds ($t_i$), for example $40 \leq t_{CPU} \leq 70$, and $t_s = 100$. Our assumption here is based on our observation that the average CPU load when the collector is working is around 40\%--hence we are using this as the lower threshold: 
\textit{$h_{r_i}^t$ = \{$u_{\text{CPU}_i}$:$t_{CPU}$, $u_{\text{RAM}_i}$:$t_{RAM}$, $u_{\text{storage}_i}$:$t_{STORAGE}$, $s_{\text{net}_i}$:$t_{s}$, $s_{\text{r}_i}$:$t_{s}$\}} is the target vector and \textit{$h_{r_i}^o$ = \{$u_{\text{CPU}_i}$:$90$, $u_{\text{RAM}_i}$:55, $u_{\text{storage}_i}$:$80$, $s_{\text{net}_i}$:$100$, $s_{\text{r}_i}$:$100$\}} is the operational vector. 

By applying the quantization ($P_{9 \rightarrow 3}$) we exclude the KPIs classified as ``normal" (i.e., $u_{\text{RAM}_i}, s_{\text{net}_i}$,$s_{\text{r}_i}$). For the rest, we apply scaling and compute the gradient, resulting with: \textit{$\nabla E$ = \{$\nabla u_{\text{CPU}_i}$:$0.57$, $\nabla u_{\text{RAM}_i}$:$0$, $\nabla u_{\text{storage}_i}$:$0.28$, $\nabla s_{\text{net}_i}$:$0$, $\nabla s_{\text{r}_i}$:$0$\}}. 
\def\thefootnote{*}\footnotetext{This logic applies to the rest of the definitions where $f$ appears. That is $f$ is a function that is specified by a policy. This allows to us to provide generic definitions, where the functions can be defined using a policy-based approach.}
The gradients indicate that both CPU and storage utilization are above the max thresholds, and as such need to be decreased to reduce the drift, and to ensure the health of the resource.

\subsection{Composite Resource Health ($k_{H_{r}}$)}
The composite resource health $k_{H_{r}}$ is evaluated similarly by applying a policy function to the vector of individual resource health KPIs ($h_{r_i}, \forall i \in \{1, 2, ..., n\}$).
\begin{equation}
    k_{H_{r}} = f(P_{9\rightarrow3}(h_{r_1}), P_{9\rightarrow3}(h_{r_2}), ..., P_{9\rightarrow3}(h_{r_n}))
\end{equation}

\subsection{Service Health ($k_{H_s}$)}
We define the health of a service as the composition of the hardware (e.g., physical, virtual), software and network that make up the service. This involves performing a heath check on the above components. If we assume composite metrics (similarly to $k_{H_{r}}$) for the software ($k_{H_{sw}}$), and the connectivity ($k_{H_{net}}$), then we can define the composite health of the service as a policy function (e.g., $min$) on the vector of these KPIs. 
\begin{equation}
    k_{H_s} = f(k_{H_{r}}, k_{H_{sw}}, k_{H_{net}}) 
\end{equation}

For a more granular analysis, we assess the KPIs related to the resources that enable one independent portion of the service (e.g., a collector and the devices for collection). Then, $h_{s_i}$ is a vector of the individual resource health ($h_{r_i}$), the software health per resource ($h_{sw_i}$), and the health of the devices we are collecting from ($h_{a_i}$), hereafter referred to as agents. As described before, a policy (or policies) will specify the required function(s). In our case, we apply the following policy function to the vector of KPIs: 
\begin{equation}
    \min_{i}(P(h_{r_i}), h_{sw_i}, h_{a_i}) 
\end{equation}
\noindent 
where the policy applies a quantization on $h_{r_i}$, and evaluates the overall health of the sub-service by taking the minimum. Here, $h_{sw_i}, h_{a_{i}}$ are binary; and $h_{a_i}$ is a vector of the agent's: resource health ($h_{r-a_{j}}$, evaluated similarly to Equation \ref{eq:resourcehealth}), software health ($h_{sw-a_{ij}}$), and network connectivity health ($h_{net_{ij}})$ between resource $i$ and $j$ (e.g., collector--agent). 
\begin{equation}
    h_{a_i} = f(h_{r-a_{j}}, h_{sw-a_{j}}, h_{net_{ij}})
\end{equation}
Some example policy functions for evaluating $h_{a_i}$ include:
\begin{itemize}
    \item a strict policy: $h_{a_i} = \min_{j}(\min(h_{r-a_{j}}, h_{sw-a_{j}}, h_{net_{ij}}))$.
    \item a relative average, where $Q$ is a quantization policy: $h_{a_i} = Q\left(\sum_{j}\min(h_{r-a_{j}}, h_{sw-a_{j}}, h_{net_{ij}})\right)$,
    \item a specific match: if $k$ agents need to be operational, where \( g \) and \( b \) are counters representing the number of agents under \textit{good} and \textit{bad} conditions, respectively, we can use the boolean expression ($B$), or relax with $g \geq 5$:
\[ B = 
\begin{cases} 
\text{true}, & \text{if } g - b \geq 5 \\
\text{false}, & \text{otherwise}
\end{cases}
\]
\end{itemize}

To assess the overall health of the service different policy functions can be used. For our use-case, we use:
\begin{equation}
    k_{H_s(\%)} = \left( \frac{\sum_{i=1}^{n} \text{normalize}(h_{s^i})}{n} \right) \times 100\%
    \label{eq:servicehealth}
\end{equation}
\begin{equation}
\text{normalize}(h{_{s_i}}) = 
\begin{cases} 
1, & \text{if } h{_{s_i}} = [0,1] \\
0, & \text{if } h{_{s_i}} = [-1] 
\end{cases}
\end{equation}

\subsection{Service Availability ($k_{A_s}$)}
To achieve the target availability, $n$ collectors are required. We use the below reliability formula to determine the number of collectors to achieve a specific availability, under the assumption that the failure of each is independent of the others.
\begin{equation}
k_{A} = 1 - \prod_{i=1}^{n}(1 - k_{A_{r_i}})
\label{eq:reliability}
\end{equation}
Where $k_{A_{r_i}}$ is a pre-defined availability per resource (in our case, $k_{A_{r_i}}=99.9\%$), and \(n\) is the number of redundant resources. We assess the availability using \(k_{A_{r_i}}=f(P(h{_{s_i}}))\), where $f$ is a function that maps the service health status:
\begin{align*}
    f(P(h{_{s_i}})) = \begin{cases} 
    100\%, & \text{if } P(h{_{s_i}}) = [1,0] \text{ (healthy)} \\
    0, & \text{if } P(h{_{s_i}}) = [-1] \text{ (unhealthy)}
    \end{cases}
\end{align*}
Let $k_{A_s}$ be the overall service availability, which depends on the health of the service, specifically the downtime. We periodically probe the service (frequency of 1min), and keep track when the service is not operational using a counter ($t_{down}$). This allows us to evaluate the service availability over the period the service should be active ($t_{planned}$): 
\begin{align}
k_{A_s} = \frac{t_{planned} - t_{down}}{t_{planned}}
\end{align}

For example, we need 2 resources ($n=2$) to provide $k_{A}=0.9999$. Then, considering we've been monitoring the service for 30days, and $t_{down}\approx2m$, then we have: $k_{A_s} = \frac{720 - 0.03}{720} \approx 0.9999$ during the 30days. Considering the whole year, to maintain this availability, we need to ensure: $t_{down} \leq 52.6m$, which is the max downtime for maintaining a service availability of 99.99\%, ($k_{A_s} = \frac{8760 - 0.8767}{8760}\approx 0.9999$).

\subsection{Intent Health}
The intent is fully met if the following condition holds, otherwise there exists some degree of drift which can be exactly determined using the previous definitions.
\begin{equation}
    \text{Intent} =
    \begin{cases}
        1, & k_{A_s} \geq 99.99\%, \\
        0, & \text{otherwise}.
    \end{cases}
\end{equation}

\section{Use-case Evaluation}\label{evaluation-use}
In this section, we evaluate the following intent: \textit{``Create collectors in Domain West for gathering Netflow data in the domain, such that the collectors have 99.99\% availability"}. First, we briefly discuss our LLM pipeline (extended with assurance and state feedback), and then describe the fulfillment and assurance steps and results, for the above intent.

\subsection{LLM Pipeline for Intent Fulfillment and Assurance}
In prior work, we introduced a pipeline of 3 consecutive LLM models (GPT), each trained via few-shot learning to perform a specific task \cite{dzeparoska2023llm}. The first LLM (\textit{classifier}) was trained to classify intents to predefined intent types (e.g., create resource, deploy service, discover resource). The output from the classifier, and the intent are then fed to the second LLM (\textit{policy generator}) which is trained to progressively decompose intents to policies using our policy model defined in \cite{dzeparoska2021towards}. Last, the third LLM (\textit{validator}) receives the policy tree, intent, and intent type, and inspects the policy tree for any omissions, incorrect sequences or attributes. In this work, we add an \textit{assurance} LLM that we teach using the assurance methodology via few-shot learning (e.g., the prompt includes the steps for assurance such as quantize, compute gradients, as well as metric and KPI definitions).

\begin{table*}[!ht]
\caption{Fulfillment and Assurance Policies generated by the LLM, with policy execution state feedback to the LLM.}
\centering
\begin{tabular}{|l|l|}
\hline
\rowcolor{mintcream}
\rule{0pt}{10pt}LLM-generated Policies for Intent Fulfillment & Feedback from policy execution \\
\hline
\rule{0pt}{8pt}M$_1$ = (get, domain, zone=West, kpi=availability) & True, \{zone: West, availability: 99.9\%\} \\
\hline
\rule{0pt}{8pt}M$_2$ = (get, switch, zone=West) & True, \{zone: West, switch:[sw$_1$, sw$_2$, sw$_3$ ...]\} \\
\hline
\rule{0pt}{8pt}A$_1$ = (compliance, domain, zone=West, availability=99.99, type=vm) & True, \{type:vm, count: 2\} \\
\hline
\rule{0pt}{8pt}A$_2$ = (avail, vm, zone=West, count=2) & True, [\{size:small, count: 50\}, \{size:medium,\\ &  count: 20\}, \{size:large, count: 15\}]  \\
\hline
\rule{0pt}{8pt}E$_1$ = (create, vm, zone=West, count=2, size=small, name=[collector$_1$, collector$_2$], image=ubuntu)
 & \scriptsize True, [\{name: collector$_1$, IP:10.0.0.10, size:small\}, \\ & 
\scriptsize \{name: collector$_2$, IP:10.0.0.11, size:small\}]  \\
 \hline
\rule{0pt}{8pt}E$_2$ = (validate, [collector$_1$,collector$_2$], zone=West)
 & \scriptsize True, [\{ssh: True, ping: True\},  \{ssh: True, ping: True\}]
  \\
  \hline
\rule{0pt}{8pt}E$_3$ = (deploy, [collector$_1$, collector$_2$], service=collector, type=netflow, name=service$_{\text{netflow}}$) & True, [\{\},\{\}]  \\
\hline
\rule{0pt}{8pt}E$_4$ = (configure, [collector$_1$, collector$_2$], service=service$_{\text{netflow}}$, source=M$_2$, zone=West) & True, [\{\},\{\}]  \\
\hline
\rule{0pt}{8pt}E$_5$ = (start, [collector$_1$, collector$_2$], service=service$_{\text{netflow}}$, zone=West) & True, [\{\},\{\}] \\
\hline
\rule{0pt}{8pt}E$_6$ = (healthcheck, service$_{\text{netflow}}$, output=App$_1$, name=health) & True, [\{\},\{\}] \\ 
\hline
\rule{0pt}{8pt}E$_7$ = (schedule, E$_6$, frequency=hourly) & True, \{\}  \\ 
\hline
\rule{0pt}{8pt}E$_8$ = (get, App$_1$, name=health, kpi=target) &  
 [\{name: service$_{\text{netflow}}$, availability:99.99\%\}, 
\\ &  \{name: service$_{\text{netflow}}$, health:100\%\}] \\ 
\hline
\rowcolor{mintcream}
\rule{0pt}{10pt}LLM-generated Policies for Intent Assurance & Feedback from policy execution \\
\hline
\rule{0pt}{8pt}E$_1$ = (get, App$_1$, name=health, kpi=operational) & [\scriptsize  \{name: service$_{\text{netflow}}$, availability:99.99\%\}, 
\\ & \scriptsize  \{name: service$_{\text{netflow}}$, health:50\%\}] \\ 
\hline
\rule{0pt}{8pt}E$_2$ = (restart, collector$_1$, zone=West) & True, \{name: collector$_1$, status: active\}  \\ 
\hline
\rule{0pt}{8pt}E$_3$ = (validate, collector$_1$, zone=West) & True, \{\}  \\ 
\hline
\rule{0pt}{8pt}E$_4$ = (start, collector$_1$, service=service$_{\text{netflow}}$, zone=West) & True, \{\}  \\ 
\hline
 \end{tabular}
 \label{table:policies}
\end{table*}

\subsection{Progressive intent decomposition guided by state input}
For each intent, we generate a set of ordered policies (a \textit{Policy Tree}) that contain actions related to a closed control loop. The use-case policies for fulfillment and assurance are provided in Table \ref{table:policies}. To build the logic progressively based on operational conditions, we feed the policy execution result back to the LLM that is trained to generate policies in sequence. To do so, we map each policy to an API and provide as feedback: a Boolean value (indicating success or failure in terms of the API call), and state information. For example, the LLM receives the following feedback upon the execution of a policy to check VM size availability: \textit{True, \{zone=Toronto, type=vm, count=[50, 20, 15], size=[small, medium, large]\}}. This allows the LLM to choose the allocation, and to enhance its reasoning, different algorithms could be provided as context information to the LLM--where the computation can be handled by the model itself, or through external solvers. 

\subsection{Use-case: fulfillment}
The LLM progressively decomposes the intent, and generates the necessary policies. Policy $A_1$ in Table \ref{table:policies} is used to determine the number of collectors required, using the parallel reliability formula given in Equation \ref{eq:reliability}. To allow the model to perform such calculations, we either need to provide insight as part of the few-shot prompt, or connect the model to a vector database for in-context learning. The latter provides more flexibility, as it can allow the model to test different solvers, and choose the appropriate solution. Given that $k_A \geq 0.9999$, and $k_{A_{r_i}}=0.999$, solving for $n$ results with $2$, that is $2$ collectors are needed in Domain West to achieve the desired combined availability of 99.99\%, ($n = \frac{\log(1 - k_A)}{\log(1 - k_{A_{r_i}})} \approx 2$). The rest of the policies then perform the necessary actions that incude creating the collectors, deploying the collection service, scheduling the health check, etc. We then regularly perform a health check, and if we detect drift, we trigger the \textit{assurance LLM}. To do so, we purposefully shut down $collector_2$.

\begin{table*}
\centering
\begin{tabular}{|l|c|c|c|c|c|c|c|c|c|c|}
\hline
\rowcolor{gray!25}
& \multicolumn{5}{c|}{\textbf{resource health $(h_{r_i}^o)$}} & \multicolumn{3}{c|}{\textbf{sub-service health $(h_{s_i}^o)$}} & \multicolumn{2}{c|}{\textbf{Corrective Actions}}\\
\cline{2-9}
\cline{10-11}
\rowcolor{gray!25} & $u_{\text{CPU}_i}$ & $u_{\text{RAM}_i}$ & $u_{\text{storage}_i}$ & $s_{\text{net}_i}$ & $s_{\text{r}_i}$ & $h_{r_i}$ & $h_{sw_i}$ & $h_{a_i}$ & $restart$ & $recreate$ \\
\hline
\rowcolor{green!7} collector$_1$ & $60$\% & $60$\% & $50$\% & $100$\% & $100$\% & $1$ & $1$ & $1$ &  n/a & n/a \\
\hline
\rowcolor{green!7} $\nabla E$ & $0$ & $0$ &$0$ & $0$ & $0$ & \multicolumn{3}{c|}{$f = min \rightarrow f(h_{s_1})=1$; $\nabla=0$} & n/a & n/a \\
\hline
\rowcolor{red!10} collector$_2$ & $0$\% & $0$\% & $0$\% & $0$\% & $0$\% & $-1$& $0$ & 0 & $\checkmark$ & \xmark   \\
\hline
\rowcolor{red!10} $\nabla E$ & $-2$ & $-2$ & $-2$ & $-2$ & $-2$ & \multicolumn{3}{c|}{$f = min \rightarrow f(h_{s_2})=-1$; $\nabla=-4$} & $P=1\times f(\nabla)$ & $P=2\times f(\nabla)$  \\
\hline
\rowcolor{gray!25}
\textbf{Service KPIs:} & \multicolumn{4}{c|}{\textbf{Target} }  &  \multicolumn{4}{c|}{\textbf{Operational at $t_1$} (drift detected)}  & \multicolumn{2}{c|}{\textbf{Operational at $t_2$} (intent assured)}\\
\hline
\rowcolor{gray!10}
\textbf{Health} & \multicolumn{4}{c|}{ $k_{H_s}^t=100\%$}  & \multicolumn{3}{c|}{$k_{H_s}^o=50\%$} & $\nabla=-1$  & $k_{H_s}^o=100\%$ & $k_{H_s}^o=100\%$   \\
\hline
\rowcolor{gray!10}
\textbf{Availability} & \multicolumn{4}{c|}{$k_{A_s}^t=0.9999$}  & \multicolumn{3}{c|}{$k_{A_s}^o=0.9999$} & $\nabla=0$ &  \textbf{$t_{down} \approx  90s$} & \textbf{$t_{down} \approx  200s$} \\
\hline
\end{tabular}
\caption{Assurance Analysis of target and operational KPIs considering two actions and their penalty impacts.}
\label{table:assurance}
\end{table*}

\subsection{Use-case: assurance}
The assurance LLM has the necessary context, i.e., the assurance methodology defined in this paper. The goal of the LLM is to review the target and operational KPIs, analyze the drift, determine the restorative actions, and generate progressively a set of assurance-related policies. To compute the above, the following data is made available to the model:
\begin{enumerate}
    \item Target data:
        \begin{itemize}
            \item $\overrightarrow{K_{T}}$ = \{$k_{A_s}$: 99.99\%\},
            \item individual resource health ($h_{r_i}^t$), in Subsection \ref{targethealth}.
            \item sub-service health: $\overrightarrow{h_{s_i}^t}$ = \{$h_{r_i}$:1, $h_{sw_i}$: 1, $h_{a_i}$:1\} 
            \item sub-service agents health: \\$\overrightarrow{h_{a_i}^t}$ = \{$h_{r_{a_j}}$:1, $h_{sw_{a_j}}$: 1, $h_{net_{a_{ij}}}$:1\}, $\forall j \in [1,m]$.
            \item composite service health: \\$\overrightarrow{k_{H_s}^t}$ = \{$k_{H_{r}}$:1, $k_{H_{sw}}$: 1, $k_{H_{net}}$:1\} $\rightarrow k^o_{H_s(\%)} = 100$.
        \end{itemize}

    \item Operational data:
    \begin{itemize}
        \item individual resource health ($h_{r_i}^o$), in Table \ref{table:assurance}.
        \item  sub-service health for collector1: \\ $\overrightarrow{h_{s_1}^o}$ = \{$h_{r_1}$:1, $h_{sw_1}$: 1, $h_{a_1}$:1\} 
        \item  sub-service health for collector2: \\ $\overrightarrow{h_{s_2}^o}$ = \{$h_{r_2}$:-1, $h_{sw_2}$: 0, $h_{a_2}$:0\} 
        \item sub-service agents health for collector1: \\$\overrightarrow{h_{a_1}^t}$ = \{$h_{r_{a_j}}$:1, $h_{sw_{a_j}}$: 1, $h_{net_{a_{1j}}}$:1\}, $\forall j \in [1,m]$.
        \item sub-service agents health for collector2: \\$\overrightarrow{h_{a_2}^t}$ = \{$h_{r_{a_j}}$:1, $h_{sw_{a_j}}$: 1, $h_{net_{a_{2j}}}$:0\}, $\forall j \in [1,m]$.
        \item composite service health: \\$\overrightarrow{k_{H_s}^o}$ = \{$k_{H_{r}}$:-1, $k_{H_{sw}}$: 0, $k_{H_{net}}$:0\} $\rightarrow k^o_{H_s(\%)} = 50$.
    \end{itemize}  
\end{enumerate}

With the above information, and using our defined methodology, the LLM applies policy functions to the above vectors (e.g., to quantize), computes the gradients, and analyzes the intent drift to recommend actions. We summarize the output in Table \ref{table:assurance} which contains the health check KPIs, policy functions applied, gradients, potential restorative actions, and action penalties (related to the time the actions take). The LLM can reason from the breakdown of the KPIs and then correctly identify the necessary action(s). That is, the model is aware that the overall service health requires addressing the issue with $collector_2$, and it connects the impact of the resource health towards the other KPIs.

As an example, we describe the general steps the LLM performs to run the health check for $collector_2$: quantize $h_{r_2}^o$ (\textit{result}: critical ($-1$)); calculate the gradient (\textit{result}: a vector of $-2$ elements). Next, assess the sub-service health $h_{s_2}^o$ by applying a policy function to the resource, software and agent health (\textit{result}: $h_{s_2}^o=-1$). Evaluate the overall service health, \textit{result}: $k^o_{H_s(\%)} = 50\%$. The LLM uses the downtime counter and determines the availability is still met, and analyzes restorative actions for $collector_2$. Given the time each action takes (reflected by the penalty factor, shown in Table \ref{table:assurance}, the LLM selects the $restart$ action). The model then waits for the execution of the policy to restart the collector ($E_2$ in Table \ref{table:policies}) and upon receiving successful feedback proceeds to generate the remaining assurance policies ($E_3$ and $E_4$) to validate and start the collection. Some of the above steps during the model's analysis are shown in Figure \ref{snippet}.

\begin{figure}
   \centering
   \includegraphics[width=\columnwidth]{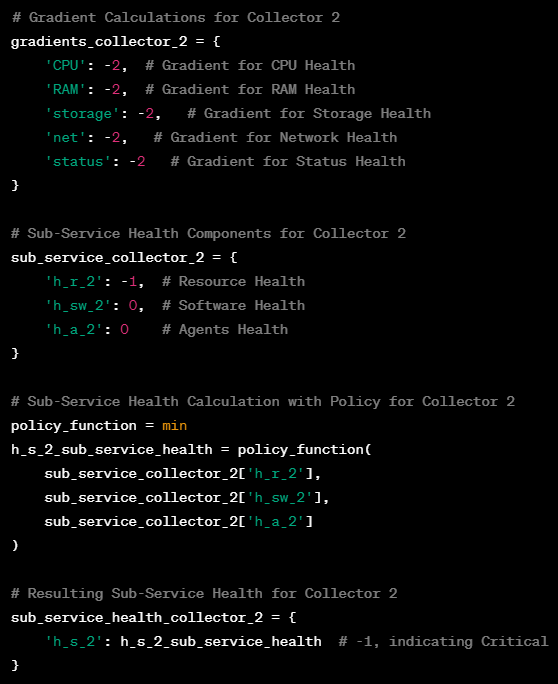} 
   \caption{Snippet of the assurance analysis by the LLM.}
   \label{snippet} 
\end{figure}

\subsection{Evaluation results: fulfillment and assurance}
We evaluate the total time to fulfill and assure our use-case intent. For the intent to policy generation we use the GPT4 API, and to deploy the intent we use our cloud testbed (SAVI). We report on the total time and provide the breakdown execution times as well. In total, we conducted 5 trials, and report on the average execution times in Table \ref{table:time}. We conclude from our experiment that we can leverage LLMs not only to understand, process and decompose intents into policies that can be mapped to API calls, but moreover to assure, and determine the necessary actions when intent drift occurs. From our evaluation, the time to generate the policies for fulfillment takes about $7$min, whereas for assurance takes about $1.5$min. Our methodology and LLM pipeline can facilitate automation, and provide quick responses that are 1-3x faster compared to manual procedures. 
\begin{table}
\caption{Intent Fulfillment and Assurance Execution Times}
\centering
\begin{tabular}{|l|c|c|}
\hline
\rowcolor{gray!20} \textbf{Policy Generation} & \textbf{Fulfillment} & \textbf{Assurance}  \\
\hline
Number of generated policies & $12$ & $4$ \\
\hline
GPT API Execution Time (s) & $23$ & $25$ \\
\hline
Cloud Testbed Execution Time (s) & $370$ & $64$ \\
\hline
\rowcolor{gray!10} Total Execition Time (s) & $393$ & $89$ \\
\hline
\end{tabular}
\label{table:time}
\end{table}

\section{Conclusions and Future Work}
We presented our assurance framework that allows us to obtain state feedback and close the loop through corrective actions. To teach the model, we leverage few-shot learning, which can make it difficult to generalize well. In future work, we plan to use vector databases for in-context learning, such that the model can quickly acquire the necessary knowledge. In addition, intent verification ensures that the policy tree results with a state close to the target one. This necessitates a digital twin, to test policies and validate the state against the intent. 

\bibliographystyle{IEEEtran}
\bibliography{IEEEabrv,references}

\begin{thebibliography}{10}
\providecommand{\url}[1]{#1}
\csname url@samestyle\endcsname
\providecommand{\newblock}{\relax}
\providecommand{\bibinfo}[2]{#2}
\providecommand{\BIBentrySTDinterwordspacing}{\spaceskip=0pt\relax}
\providecommand{\BIBentryALTinterwordstretchfactor}{4}
\providecommand{\BIBentryALTinterwordspacing}{\spaceskip=\fontdimen2\font plus
\BIBentryALTinterwordstretchfactor\fontdimen3\font minus \fontdimen4\font\relax}
\providecommand{\BIBforeignlanguage}[2]{{%
\expandafter\ifx\csname l@#1\endcsname\relax
\typeout{** WARNING: IEEEtran.bst: No hyphenation pattern has been}%
\typeout{** loaded for the language `#1'. Using the pattern for}%
\typeout{** the default language instead.}%
\else
\language=\csname l@#1\endcsname
\fi
#2}}
\providecommand{\BIBdecl}{\relax}
\BIBdecl

\bibitem{rfc9315}
\BIBentryALTinterwordspacing
A.~Clemm, L.~Ciavaglia, L.~Z. Granville, and J.~Tantsura, ``{Intent-Based Networking - Concepts and Definitions},'' RFC 9315, Oct. 2022. [Online]. Available: \url{https://www.rfc-editor.org/info/rfc9315}
\BIBentrySTDinterwordspacing

\bibitem{martini2023intent}
B.~Martini, M.~Gharbaoui, and P.~Castoldi, ``Intent-based network slicing for sdn vertical services with assurance: Context, design and preliminary experiments,'' \emph{Future Generation Computer Systems}, vol. 142, pp. 101--116, 2023.

\bibitem{zheng2021network}
X.~Zheng and A.~Leivadeas, ``Network assurance in intent-based networking data centers with machine learning techniques,'' in \emph{2021 17th International Conference on Network and Service Management (CNSM)}.\hskip 1em plus 0.5em minus 0.4em\relax IEEE, 2021, pp. 14--20.

\bibitem{falkner2022intent}
M.~Falkner and J.~Apostolopoulos, ``Intent-based networking for the enterprise: a modern network architecture,'' \emph{Communications of the ACM}, vol.~65, no.~11, pp. 108--117, 2022.

\bibitem{samdanis2023ai}
K.~Samdanis, A.~N. Abbou, J.~Song, and T.~Taleb, ``Ai/ml service enablers \& model maintenance for beyond 5g networks,'' \emph{IEEE Network}, 2023.

\bibitem{perepu2022intent}
S.~K. Perepu, J.~P. Martins, R.~Souza, and K.~Dey, ``Intent-based multi-agent reinforcement learning for service assurance in cellular networks,'' in \emph{GLOBECOM 2022-2022 IEEE Global Communications Conference}.\hskip 1em plus 0.5em minus 0.4em\relax IEEE, 2022, pp. 2879--2884.

\bibitem{wang2023artificial}
J.~Wang, J.~Liu, J.~Li, and N.~Kato, ``Artificial intelligence-assisted network slicing: Network assurance and service provisioning in 6g,'' \emph{IEEE Vehicular Technology Magazine}, vol.~18, no.~1, pp. 49--58, 2023.

\bibitem{de2022machine}
N.~F.~S. de~Sousa, M.~T. Islam, R.~U. Mustafa, D.~A.~L. Perez, C.~E. Rothenberg, and P.~H. Gomes, ``Machine learning-assisted closed-control loops for beyond 5g multi-domain zero-touch networks,'' \emph{Journal of Network and Systems Management}, vol.~30, no.~3, p.~46, 2022.

\bibitem{gallego2022machine}
J.~Gallego-Madrid, R.~Sanchez-Iborra, P.~M. Ruiz, and A.~F. Skarmeta, ``Machine learning-based zero-touch network and service management: A survey,'' \emph{Digital Communications and Networks}, vol.~8, no.~2, pp. 105--123, 2022.

\bibitem{perepu2022multi}
S.~K. Perepu, J.~P. Martins, K.~Dey \emph{et~al.}, ``Multi-agent reinforcement learning for intent-based service assurance in cellular networks,'' \emph{arXiv preprint arXiv:2208.03740}, 2022.

\bibitem{mohamed2023automatic}
R.~Mohamed, I.~Lambadaris, A.~Leivadeas, J.~Chinneck, T.~Morris, and P.~Djukic, ``Automatic feasibility restoration for 5g cloud gaming,'' in \emph{ICC 2023-IEEE International Conference on Communications}.\hskip 1em plus 0.5em minus 0.4em\relax IEEE, 2023, pp. 846--851.

\bibitem{mai2022end}
V.~S. Mai, R.~J. La, T.~Zhang, and A.~Battou, ``End-to-end quality-of-service assurance with autonomous systems: 5g/6g case study,'' in \emph{2022 IEEE 19th Annual Consumer Communications \& Networking Conference (CCNC)}.\hskip 1em plus 0.5em minus 0.4em\relax IEEE, 2022, pp. 644--651.

\bibitem{9844116}
K.~Edeline, T.~Carlisi, J.~Iurman, B.~Claise, and B.~Donnet, ``Towards a closed-looped automation for service assurance with the dxagent,'' in \emph{2022 IEEE 8th International Conference on Network Softwarization (NetSoft)}, 2022, pp. 67--72.

\bibitem{leivadeas2022survey}
A.~Leivadeas and M.~Falkner, ``A survey on intent based networking,'' \emph{IEEE Communications Surveys \& Tutorials}, 2022.

\bibitem{leivadeas2023autonomous}
A.~Leivadeas and F.~Matthias, ``Autonomous network assurance in intent based networking: Vision and challenges,'' pp. 1--10, 2023.

\bibitem{10418089}
K.~Dzeparoska, A.~Tizghadam, and A.~Leon-Garcia, ``Emergence: An intent fulfillment system,'' \emph{IEEE Communications Magazine}, pp. 1--6, 2024.

\bibitem{jacobs2021hey}
A.~S. Jacobs, R.~J. Pfitscher, R.~H. Ribeiro, R.~A. Ferreira, L.~Z. Granville, W.~Willinger, and S.~G. Rao, ``Hey, lumi! using natural language for $\{$intent-based$\}$ network management,'' in \emph{2021 USENIX Annual Technical Conference (USENIX ATC 21)}, 2021, pp. 625--639.

\bibitem{9687516}
Y.~Ouyang, C.~Yang, Y.~Song, X.~Mi, and M.~Guizani, ``A brief survey and implementation on refinement for intent-driven networking,'' \emph{IEEE Network}, vol.~35, no.~6, pp. 75--83, 2021.

\bibitem{kleene1938notation}
S.~C. Kleene, ``On notation for ordinal numbers,'' \emph{The Journal of Symbolic Logic}, vol.~3, no.~4, pp. 150--155, 1938.

\bibitem{dzeparoska2023llm}
K.~Dzeparoska, J.~Lin, A.~Tizghadam, and A.~Leon-Garcia, ``Llm-based policy generation for intent-based management of applications,'' in \emph{2023 19th International Conference on Network and Service Management (CNSM)}.\hskip 1em plus 0.5em minus 0.4em\relax IEEE, 2023, pp. 1--7.

\bibitem{dzeparoska2021towards}
K.~Dzeparoska, N.~Beigi-Mohammadi, A.~Tizghadam, and A.~Leon-Garcia, ``Towards a self-driving management system for the automated realization of intents,'' \emph{IEEE Access}, vol.~9, pp. 159\,882--159\,907, 2021.

\end{thebibliography}
\end{document}